\documentclass{article}

\usepackage{arxiv_preprint,times}
\usepackage[T1]{fontenc}
\usepackage[utf8]{inputenc}
\usepackage{amsmath,amssymb}
\usepackage{graphicx}
\usepackage{booktabs}
\usepackage{longtable}
\usepackage{float}
\usepackage{hyperref}
\usepackage{url}
\usepackage{placeins}

\hypersetup{
  pdftitle={Learning Functional Specialisation for Bimanual Robotic Control},
  pdfauthor={Jingwen Li, Levin Kuhlmann, Jason Friedman, and Gideon Kowadlo}}

\title{Learning Functional Specialisation\\for Bimanual Robotic Control}
\author{
\normalfont
\begin{tabular}{c}
\textbf{Jingwen Li$^{1}$, Levin Kuhlmann$^{1}$, Jason Friedman$^{2}$, and Gideon Kowadlo$^{1,3}$}\\[0.8em]
$^{1}$Department of Data Science and AI, Monash University\\[2pt]
$^{2}$Department of Physical Therapy \& Sagol School of Neuroscience, Tel Aviv University\\[2pt]
$^{3}$Cerenaut\\[0.45em]
\texttt{jlii0556@student.monash.edu, levin@monash.edu.au}\\
\texttt{jason@tau.ac.il, gideon@cerenaut.ai}
\end{tabular}}
\arxivfinalcopy

\begin{document}
\maketitle
\lhead{}

\begin{abstract}
Bimanual robotic control often requires diverse functional demands to be handled
simultaneously under challenging control conditions. Modular controllers provide one way to distribute these demands, but how control responsibilities should be organised across controllers and the two arms, and how this organisation affects performance and flexibility, remains unclear. Biological motor control provides an organisational principle in which each hemisphere predominantly controls the contralateral limb, alongside functional specialisation and stable hand preferences. We therefore investigate whether analogous organisational principles can benefit modular robotic motor control in a musculoskeletal system with two communicating recurrent controllers. Using a $2\times2$ design, we compare differential (controller-specific) and shared functional objectives together with fixed and balanced (role-varying) hand--task assignments. We find that task-aligned functional specialisation is strongest when differential objectives are combined with fixed hand--task assignment aligned with contralateral control routing. This condition achieves the highest performance and control efficiency on the task, reaching 81.33\% bimanual success with significant improvements over the other training conditions. Its performance advantage over shared objectives increases under stronger motor noise but narrows under stronger energy penalties. It also shows lower hand-role flexibility: reversing the hand roles decreases bimanual success by 80.46 percentage points under differential objectives with fixed hand--task assignment, compared with only 1.51 percentage points under shared objectives with balanced hand--task assignment. Our findings show that, in bimanual robotic control with each controller more strongly routed to one hand, functional specialisation can develop through the organisation of learning responsibility and is associated with advantages in performance and control efficiency, whereas more distributed training objectives and task assignments favour greater interchangeability when roles change.
\end{abstract}

\section{Introduction}
\label{sec:introduction}
Bimanual behaviour often relies on a division of labour between the two hands rather than identical contributions \citep{swinnenIntermanualCoordinationBehavioural2002}. In bimanual robotic manipulation, the two arms often take complementary roles, with one arm providing stabilisation while the other performs the primary manipulation \citep{liuVoxActBVoxelBased2024,yuManiGaussianGeneralRobotic2025}. In human bimanual control, handedness is associated with stable preferences for using particular hands for particular motor functions \citep{woytowiczHandednessResultsComplementary2018a}. Such stable hand--function associations provide repeated experience in the same role, reinforcing learned advantages for recurring control demands \citep{provinsSpecificityMotorSkill1997,marcoriChangingHandednessWhat2019}. At the neural level, bimanual control is also organised asymmetrically: the two cerebral hemispheres make different task-dependent contributions to motor control, with each hemisphere predominantly controlling the contralateral limb while retaining weaker ipsilateral influence \citep{serrienDynamicsHemisphericSpecialization2006a,brinkmanCerebralControlContralateral1973}. This organisation establishes a stable hemisphere--limb correspondence while allowing each hemisphere to influence both limbs. The two hemisphere--limb systems also exhibit complementary functional specialisations: the dominant system has been associated with movement dynamics and trajectory control, whereas the non-dominant system shows advantages in positional stabilisation \citep{sainburgHandednessDifferentialSpecializations2005a,muthaEffectsBrainLateralization2012a}. Such functional specialisation supports a division of labour in which different systems contribute according to their respective strengths, allowing complementary control demands to be handled by different parts of the system instead of requiring equal involvement in every aspect of the task.

Modular neural controllers can support functional differentiation, but
modularity alone does not determine how responsibilities should be distributed
\citep{wolpertMultiplePairedForward1998,harunoMOSAICModelSensorimotor2001,
michaelsGoaldrivenModularNeural2020}.
Different training objectives encourage components to develop distinct functional priorities \citep{dobsBrainlikeFunctionalSpecialization2022a,rinaldoLeftRightBrain2025}, while repeated practice reinforces hand-specific motor competencies
\citep{marcoriChangingHandednessWhat2019,dexheimerRolesHandednessHemispheric2024,aracArmDominanceEmerges}.
These mechanisms have largely been studied separately, leaving unclear how these two forms of organisation interact during learning and when the resulting specialisation improves behaviour.

Theoretical accounts suggest that the benefits of specialised organisation are expressed differently under different control pressures  
\citep{seoaneOptimalityPressuresLateralization2023a}. In motor control,
signal-dependent noise causes execution variability to increase with
motor-command magnitude \citep{harrisSignaldependentNoiseDetermines1998a}, while energetic cost creates pressure to achieve task goals with lower motor expenditure \citep{todorovOptimalFeedbackControl2002a}. Interhemispheric
communication is also delayed: transcallosal conduction between motor cortical
areas has been estimated at approximately 13 ms
\citep{meyerInhibitoryExcitatoryInterhemispheric1995a}, and precisely timed
interhemispheric interactions contribute to successful bimanual coordination
\citep{liuzziCoordinationUncoupledBimanual2011}. These considerations motivate testing how motor noise, energetic cost, and inter-controller communication delay influence the behavioural benefits of learned functional specialisation.

Taken together, these observations motivate us to incorporate these organisational principles into bimanual robotic motor control. We study how controller-level functional objectives and stable hand--task assignment interact to produce functional specialisation, and how the resulting organisation affects performance, efficiency, and hand-role flexibility under control and task conditions. To do so, we use a bilateral recurrent controller in which each controller primarily controls the contralateral hand while retaining weaker influence over the ipsilateral hand. This fixed controller--hand mapping provides a stable foundation for examining how controller-level functional responsibility interacts with hand-role assignment. The controller operates in a musculoskeletal bimanual task combining trajectory tracking and positional stabilisation. Our main contributions are as follows:
\begin{itemize}
\item We introduce a controlled $2\times2$ framework that disentangles
controller-level functional objectives, defined as either separate optimisation
of the two controllers for trajectory tracking and positional stabilisation or a
shared objective combining these two functional demands, from hand--task assignment within a bimanual task where one
hand tracks a moving target while the other stabilises against external
perturbations.

\item We show that specialisation is strongest when controller-specific objectives are combined with fixed hand--task assignment. This organisation is accompanied by higher bimanual performance and efficiency, with the benefit of controller-specific objectives increasing under stronger motor noise and decreasing under stronger energy penalties. For robotic motor control, these results suggest that functional responsibility across modular controllers should be considered together with effector roles rather than designed independently.

\item We show that stronger functional specialisation can be associated with
reduced hand-role flexibility when functional demands are reassigned across
hands. Controllers trained with shared functional objectives and varied
hand--task assignments perform better when hand roles change, revealing different balances between role-specific performance and flexibility across changing bimanual task configurations.
\end{itemize}

\section{Related work}
\label{sec:related}
\paragraph{Modularity and specialisation in robot learning.} 
Computational models have examined functional and hemispheric specialisation across cognitive domains, including language and visual processing \citep{changUnifiedNeurocomputationalBilateral2020a,behrmannEmergenceTopographyHemispheric2026,dobsBrainlikeFunctionalSpecialization2022a}. In robot learning, modular architectures have been used to distribute task or functional demands across specialised components, including separate motor-control policies, task-specific modules, and expert-based manipulation policies \citep{huModularNeuralNetwork2023,jianPolicyStitching2023,liMorphologicallySymmetric2025,dengSemanticallyStructured2026}. Functional differentiation has also been studied in bilateral motor control, where controllers trained with different functional objectives can develop complementary preferences for reaching and positional stabilisation \citep{rinaldoLeftRightBrain2025}. Related work has examined how communication contributes to performance in bilateral modular controllers under control constraints \citep{liWhenDoesCommunication2026}. These studies do not establish whether controller-specific objectives alone are sufficient, or how they interact with stable hand--task experience to produce specialisation and affect motor-control performance, efficiency, and flexibility.
\vspace{-0.2em}

\paragraph{Learning responsibility across multiple objectives.}

Multi-task learning can improve generalisation by sharing information across
related tasks, but optimising multiple tasks together can also produce negative transfer when task objectives compete
\citep{zhangSurveyMultiTask2022,navonMultiTaskLearning2022,banFairResource2024}.
Existing approaches address this tension through task-specific components and
explicit task routing \citep{dingMitigatingTaskInterference2023}, specialised
expert allocation
\citep{chenModSquad2023,shazeer2017,fedusSwitchTransformersScaling2021}, or
optimisation methods that balance competing task losses
\citep{liuFAMO2023}. These approaches primarily determine how learning signals
and model capacity are distributed across tasks or model components, without
separately considering whether the same functional demands are repeatedly
associated with particular effectors during training. It therefore remains
unclear whether learned functional specialisation arises from
controller-specific objectives, repeated hand--function experience, or their
interaction. Our work separates these two sources of learning responsibility
within the same bilateral control architecture.
\vspace{-0.2em}

\section{Learning setup and measurements}
\label{sec:methods}

\subsection{Simulated Musculoskeletal Bimanual System and Tasks}
\label{sec:task}

\begin{figure}[htbp]
  \centering
  \includegraphics[width=\linewidth]{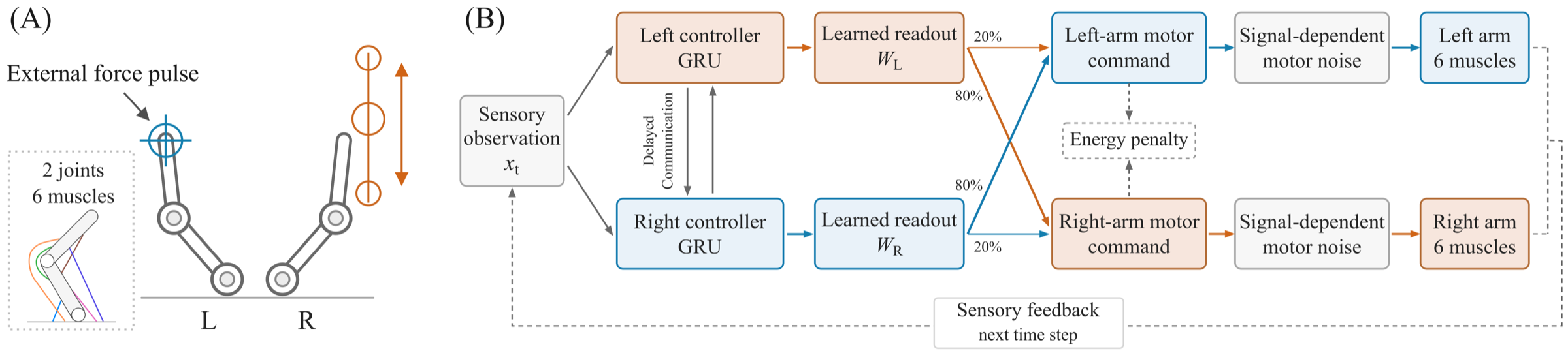}
  \caption{\textbf{(A)} Bimanual task and musculoskeletal effectors. One hand stabilises a fixed target against
  external force perturbations (left) while the other tracks a
  continuously moving target (right). The inset shows the MotorNet RigidTendonArm26 effector used for each arm, a planar two-degree-of-freedom model actuated by six Hill-type muscles.
  Two identical effectors are arranged symmetrically in the workspace. \textbf{(B)} Bilateral modular controller architecture. Two GRU-based controllers receive shared sensory observations, exchange information through delayed bidirectional connections, and contribute motor commands to the two arms. Motor commands are subject to signal-dependent noise, with motor-command magnitude determining the energetic penalty during training.}
  \label{fig:musculoskeletal-bimanual-task}
\end{figure}

We constructed a simulated bimanual control system in MotorNet
\citep{codolMotorNetPythonToolbox2024a} using two RigidTendonArm26 musculoskeletal effectors
(Fig.~\ref{fig:musculoskeletal-bimanual-task}A). The two arms were controlled simultaneously and were not mechanically coupled. The bimanual task comprised two simultaneous task components: holding and
tracking. In the holding task, one hand maintained a fixed target while resisting external
force perturbations with magnitudes sampled from 0.5--1.5~N. In the tracking
task, the other hand followed a continuously moving target with frequency
sampled from 0.75--1.25~Hz (Fig.~\ref{fig:musculoskeletal-bimanual-task}A).
Further environment and task details are provided in Appendix~\ref{app:task}.

\vspace{-0.2em}

\subsection{Bilateral Modular Controller Architecture}
\label{sec:objectives}
Inspired by human motor control, in which the two hemispheres interact during bimanual coordination
\citep{serrienDynamicsHemisphericSpecialization2006a,pauwelsRoleCorpusCallosum2023},
we designed a bilateral controller comprising two recurrent modules representing interacting hemispheric controllers
(Fig.~\ref{fig:musculoskeletal-bimanual-task}B). Each module was implemented as
a gated recurrent unit (GRU) \citep{choLearningPhraseRepresentations2014} with
64 hidden units and received the full 46-dimensional sensory observation of the
bimanual system. The controllers maintained separate recurrent states and exchanged learned signals bidirectionally with communication delay \(d\).

Each controller also projected its hidden state through a learned motor readout to produce commands for all twelve muscle actuators. To reflect the predominance of contralateral motor control together with weaker ipsilateral contributions \citep{brinkmanCerebralControlContralateral1973}, the two controller outputs were combined using fixed 80:20 contralateral-to-ipsilateral routing at the motor-readout stage, before the controller contributions were summed and transformed into muscle commands. This routing structure was held fixed across all training conditions and therefore acted as a common architectural prior, allowing both controllers to contribute to both arms while assigning stronger gains to contralateral outputs. Further architecture and control-constraint details are provided in Appendix~\ref{app:architecture}.

\subsection{Learning Responsibility Design}
\label{sec:responsibility}

\begin{figure}[h]
  \centering
  \includegraphics[width=0.9\linewidth]{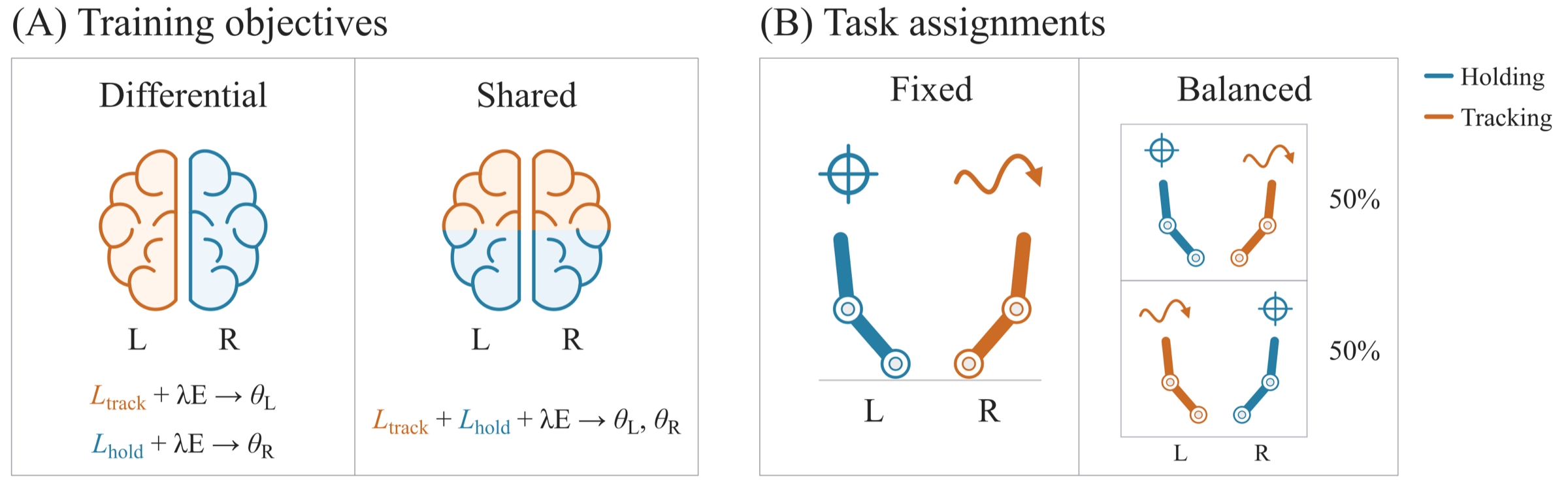}
  \caption{Experimental design for learning responsibility.
  \textbf{(A)} Training-objective structure. Differential training routes tracking and holding losses separately to the two controllers, whereas Shared training applies the combined functional loss to both controllers, with the energy term common to both schemes.
  \textbf{(B)} Task-assignment structure. Fixed training maintains the same
  hand--task mapping throughout training, whereas balanced training presents
  right-tracking/left-holding and left-tracking/right-holding mappings equally
  across trials.}
  \label{fig:learning-responsibility-design}
\end{figure}

We varied learning responsibility across two dimensions
(Fig.~\ref{fig:learning-responsibility-design}):
\textbf{(A) training objectives}, which determine which functional losses
update each controller, and \textbf{(B) task assignments}, which determine the
hand--function mappings presented during training. The forward control
architecture remained unchanged across all conditions.

\textbf{Training objectives.}
Let $\mathcal{L}_{\mathrm{track}}$, $\mathcal{L}_{\mathrm{hold}}$, and
$\mathcal{L}_{\mathrm{energy}}$ denote the tracking, holding, and energetic
losses. Both objective schemes included the same energetic loss, weighted by
$\lambda$. Under Differential training, the tracking and holding objectives
were assigned to the left and right controller parameter groups, respectively,
with gradients restricted to the corresponding group (Fig.~\ref{fig:learning-responsibility-design}A):
\begin{align}
\mathcal{J}_{L}
&= \mathcal{L}_{\mathrm{track}}
+ \lambda\mathcal{L}_{\mathrm{energy}}, \\
\mathcal{J}_{R}
&= \mathcal{L}_{\mathrm{hold}}
+ \lambda\mathcal{L}_{\mathrm{energy}}.
\end{align}
Under Shared training, both task losses were combined in a single objective
whose gradients updated all trainable parameters:
\begin{equation}
\mathcal{J}_{\mathrm{shared}}
= \mathcal{L}_{\mathrm{track}}
+ \mathcal{L}_{\mathrm{hold}}
+ \lambda\mathcal{L}_{\mathrm{energy}}.
\end{equation}

\textbf{Task assignment.} Fixed training maintained the
right-tracking/left-holding canonical mapping, whereas
Balanced training presented the canonical and reversed
left-tracking/right-holding mappings equally across trials
(Fig.~\ref{fig:learning-responsibility-design}B).

Crossing these factors produced four conditions: Differential-Fixed,
Differential-Balanced, Shared-Fixed, and Shared-Balanced. Further objective and gradient-assignment details are in Appendix~\ref{app:objectives}.

\subsection{Control constraints}
To evaluate these learning-responsibility schemes under control constraints, we tested three biologically motivated factors:
\textbf{(1) Energy penalty
weight $\lambda$} controlled the contribution of the motor-output penalty during
training. Because generating motor commands incurs
energetic cost, $\lambda$ determines how strongly energy use is weighted relative
to task performance
\citep{todorovOptimalFeedbackControl2002a}. We used $\lambda=0$ as a no-energy-penalty baseline and logarithmically spaced non-zero values $\{10^{-5},10^{-4},10^{-3}\}$ for progressively stronger relative penalties.
\textbf{(2) Motor-noise gain $k$} controlled signal-dependent perturbations applied to motor commands, with perturbation amplitude proportional to the motor command and scaled by $k$ \citep{harrisSignaldependentNoiseDetermines1998a,hamiltonControllingStatisticsAction2002}. Higher $k$ therefore increases execution noise and reduces command-to-output reliability. This tests unreliable motor execution, relevant to degraded motor function and robotic component degradation. We included $k=0$ as a deterministic baseline and evaluated stronger noise levels at $k \in \{0.05,0.1,0.2\}$.
\textbf{(3) Communication delay $d$} determined the delay applied to
bidirectional inter-controller signals, motivated by transcallosal
conduction delays and the temporal requirements of interhemispheric
coordination
\citep{meyerInhibitoryExcitatoryInterhemispheric1995a,
liuzziCoordinationUncoupledBimanual2011}, and was evaluated at
$d \in \{0,5,10,20\}$ simulation steps (50, 100, and 200 ms). These factors formed a full factorial grid of 64 control conditions. Implementation details are provided in Appendix~\ref{app:architecture}.

\vspace{-0.2em}
\subsection{Experimental Protocol}
\label{sec:constraints}
\vspace{-0.2em}
\paragraph{Training protocol}
Each control condition was trained with five random seeds. Models were
optimised using Adam \citep{kingmaAdamMethod2015} with a learning
rate of $10^{-3}$ for at most 2,000 epochs. Training details are provided in Appendix~\ref{app:training-protocol}.
\vspace{-0.2em}
\paragraph{Evaluation metrics}
Functional organisation was characterised using unit-level task preferences measured under tracking and holding demands, summarised using two complementary measures:
\textbf{(1) signed role alignment}, which measures whether hidden-unit task preferences are relatively more tracking-oriented in the left controller and more holding-oriented in the right controller, and
\textbf{(2) functional separation}, defined by the Wasserstein distance between
the two controller task-preference distributions. Behavioural performance was measured by \textbf{bimanual success rate (Bimanual SR)}, \textbf{tracking success rate (Tracking SR)}, and \textbf{holding success rate (Holding SR)}, and control efficiency by \textbf{mean control energy} and \textbf{energy per success}. Definitions are in Appendix~\ref{app:evaluation-metrics}.
\vspace{-0.8em}
\section{Results}
\label{sec:results}
\vspace{-0.2em}
\subsection{TASK ASSIGNMENT AND TRAINING OBJECTIVES INTERACT TO STRENGTHEN SPECIALISATION}
\label{sec:organisation}

Functional specialisation was strongest when controller-specific objectives
were combined with fixed hand--task assignment. Across both organisation
measures, either factor alone produced weaker organisation, indicating an
interaction between objective structure and task assignment.

\begin{figure}[h]
  \centering
  \includegraphics[width=\linewidth]{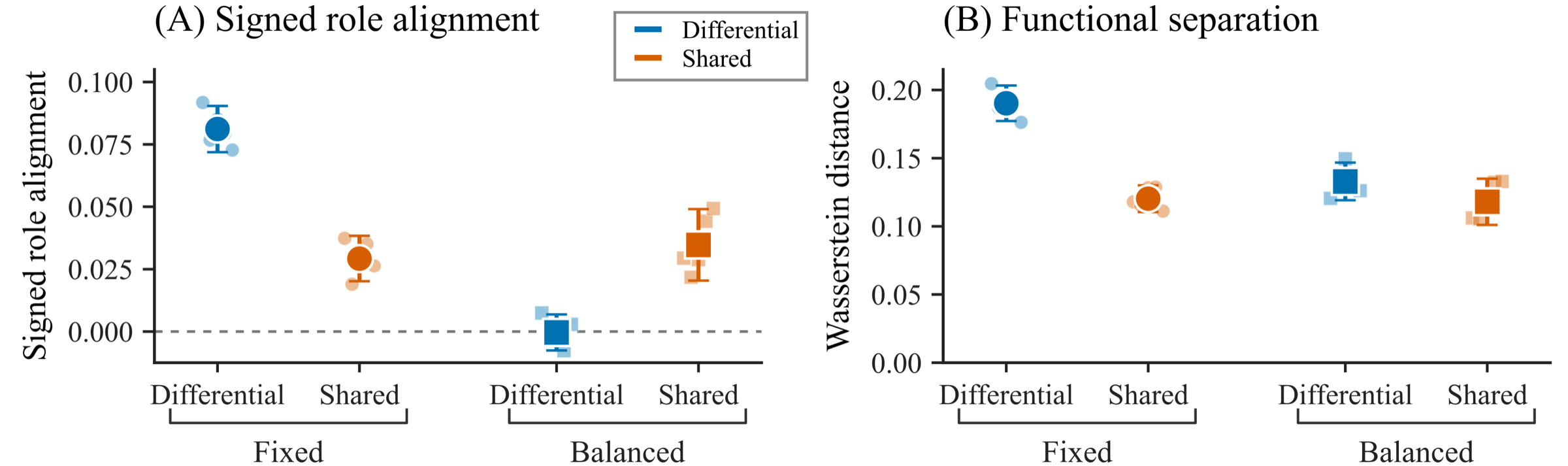}
  \caption{Functional specialisation across task assignments and training objectives. \textbf{(A)} Role alignment across training conditions, where larger positive values indicate stronger alignment with the assigned tracking and holding roles, near-zero values indicate little role-aligned preference, and negative values indicate reversed alignment. \textbf{(B)} Functional separation across training conditions, where larger values indicate greater separation between controller task-preference distributions. Large markers show five-seed means. Small markers show seed means across 64 control conditions, with horizontal jitter for visibility. Error bars denote 95\% Student's $t$ confidence intervals across seeds.}

  \label{fig:organisation}
\end{figure}

Signed role alignment was highest under Differential-Fixed training (0.081),
compared with 0.029 for Shared-Fixed, $-0.0004$ for Differential-Balanced, and
0.035 for Shared-Balanced (Fig.~\ref{fig:organisation}A).
Differential-Balanced therefore showed almost no task-aligned preference despite
retaining controller-specific objectives. Functional separation showed the same overall pattern
(Fig.~\ref{fig:organisation}B). Differential-Fixed produced the
largest separation (0.190), compared with 0.120 for Shared-Fixed, 0.133 for
Differential-Balanced, and 0.118 for Shared-Balanced.

To test whether the effect of training objectives depended on task assignment,
we computed an objective-by-assignment interaction contrast within each seed
after averaging over the 64 control conditions defined in Section~\ref{sec:constraints}. The interaction was
positive for both signed role alignment and functional separation (significant at $p<0.05$). Thus, differential objectives strengthened functional
specialisation more under fixed than balanced hand--task assignment, across both organisation measures. Full-grid results are provided
in Appendix~\ref{app:specialisation-results}.
\vspace{-0.2em}

\subsection{Specialisation Accompanies Higher Performance and Control Efficiency}
\label{sec:utility}

Differential-Fixed achieved the highest Bimanual SR (81.33\%), exceeding all alternatives. The higher Bimanual SR of Differential-Fixed was reflected in both task components, with the highest tracking and holding success rates (92.31\% and 87.08\%) and the lowest errors (Table~\ref{tab:bimanual-performance-efficiency}).
\vspace{-0.2em}
\begin{table}[h]
  \centering
  \caption{Bimanual performance and control efficiency under the right-tracking/left-holding mapping. Values are mean $\pm$ standard deviation across five means, each averaged over 64 control conditions defined in Section~\ref{sec:constraints}. $^{*}$ denotes a significant difference from Differential-Fixed after Holm correction ($p_{\mathrm{Holm}}<0.05$). Arrows indicate the preferred direction of each metric.}
  \label{tab:bimanual-performance-efficiency}
  \resizebox{\linewidth}{!}{%
    \begin{tabular}{@{}lrrrr@{}}
      \toprule
      Metric & Differential-Fixed & Shared-Fixed &
      Differential-Balanced & Shared-Balanced \\
      \midrule
      \textbf{Bimanual SR $\uparrow$} &
      \textbf{81.33 $\pm$ 3.93\%} & 71.10 $\pm$ 6.69\%$^{*}$ & 28.03 $\pm$ 7.45\%$^{*}$ & 41.36 $\pm$ 4.92\%$^{*}$ \\
      Tracking SR $\uparrow$ &
      \textbf{92.31 $\pm$ 1.88\%} & 89.28 $\pm$ 2.35\%$^{*}$ & 64.33 $\pm$ 4.17\%$^{*}$ & 84.38 $\pm$ 2.16\%$^{*}$ \\
      Holding SR $\uparrow$ &
      \textbf{87.08 $\pm$ 4.67\%} & 76.77 $\pm$ 7.81\%$^{*}$ & 45.05 $\pm$ 10.03\%$^{*}$ & 46.82 $\pm$ 5.69\%$^{*}$ \\
      Trajectory error (mm) $\downarrow$ &
      \textbf{13.45 $\pm$ 0.60} & 14.00 $\pm$ 0.42$^{*}$ & 20.29 $\pm$ 2.70$^{*}$ & 15.12 $\pm$ 0.42$^{*}$ \\
      Holding error (mm) $\downarrow$ &
      \textbf{6.79 $\pm$ 0.60} & 7.99 $\pm$ 0.63$^{*}$ & 11.63 $\pm$ 0.80$^{*}$ & 10.81 $\pm$ 0.34$^{*}$ \\
      Mean control energy $\downarrow$ &
      0.550 $\pm$ 0.135 & 0.683 $\pm$ 0.171$^{*}$ & \textbf{0.510 $\pm$ 0.093} & 0.647 $\pm$ 0.120$^{*}$ \\
      \textbf{Energy per success $\downarrow$} &
      \textbf{0.679 $\pm$ 0.172} & 0.975 $\pm$ 0.291$^{*}$ & 1.955 $\pm$ 0.712$^{*}$ & 1.558 $\pm$ 0.168$^{*}$ \\
      \bottomrule
    \end{tabular}%
  }
\end{table}

Differential-Fixed also maintained relatively low control energy (0.550), below both Shared conditions. Differential-Balanced used slightly less energy (0.510), although this difference was not significant and was accompanied by substantially lower success. When control expenditure was considered relative to successful behaviour, Differential-Fixed achieved the lowest energy per success (0.679), indicating the highest control efficiency among the four training conditions.

\begin{figure}[h]
  \centering
  \includegraphics[width=\linewidth]{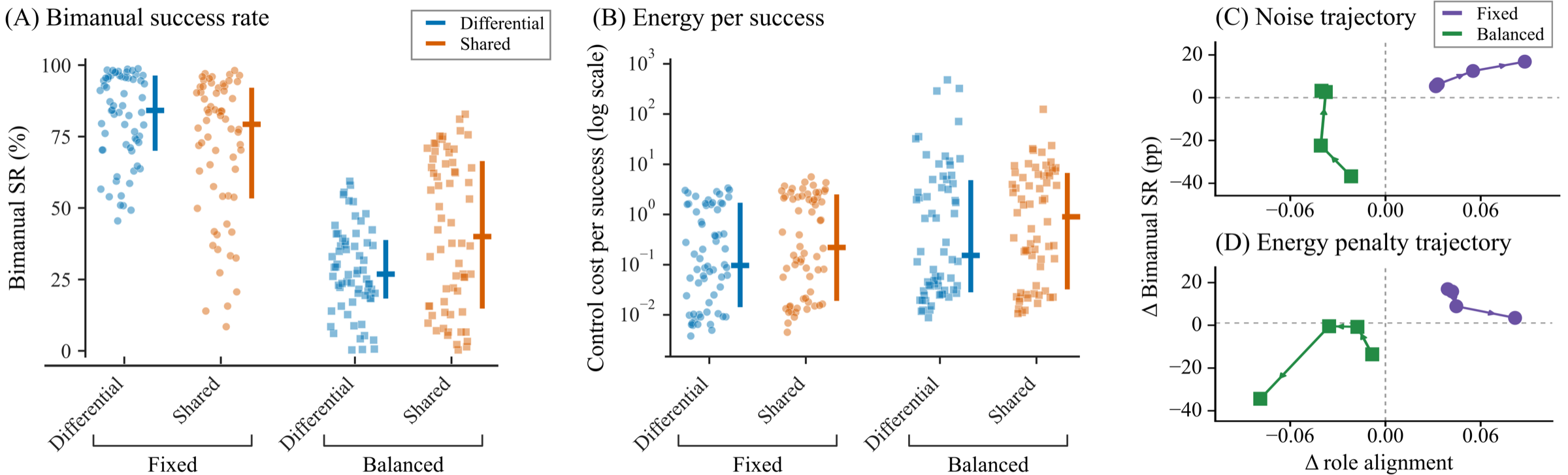}
  \caption{Bimanual performance and control efficiency across control conditions.
  \textbf{(A)} Bimanual SR under the canonical mapping.
  \textbf{(B)} Energy per success (log scale). Points show five-seed means for
  each of 64 control conditions. Horizontal and vertical bars indicate the
  median and interquartile range across conditions.
  \textbf{(C--D)} Differential--Shared differences in signed role alignment and
  Bimanual SR as motor-noise gain or energy
  penalty increases, averaging over the remaining factors. Positive $\Delta$
  values indicate higher values under Differential training.}
  \label{fig:performance-efficiency-conditions}
\end{figure}

Across the 64 control conditions, Differential-Fixed also showed the highest
median Bimanual SR and the lowest median energy per success, indicating that
its performance and efficiency advantages were broadly reflected across the
control-condition grid (Fig.~\ref{fig:performance-efficiency-conditions}A--B).

At the same time, behavioural gains were not proportional to specialisation strength
(Fig.~\ref{fig:performance-efficiency-conditions}C--D). Along the motor-noise
trajectory, larger Differential--Shared differences in signed role alignment were
accompanied by larger Bimanual SR advantages under fixed assignment. Along the
energy-penalty trajectory, the Bimanual SR advantage instead declined towards
zero while the role-alignment difference remained positive. Taken together, the condition with the strongest task-aligned functional specialisation also achieved the highest performance and best control efficiency, but stronger specialisation did not consistently correspond to larger behavioural gains across control conditions. Full-grid results and statistical comparisons are provided in
Appendices~\ref{app:specialisation-results} and
\ref{app:performance-efficiency-results}.

\vspace{-0.2em}

\FloatBarrier
\subsection{Differential Training Objectives Become More Advantageous Under Higher Motor Noise}
\label{sec:motor-noise}
\vspace{-0.2em}

\begin{figure}[h]
  \centering
  \includegraphics[width=\linewidth]{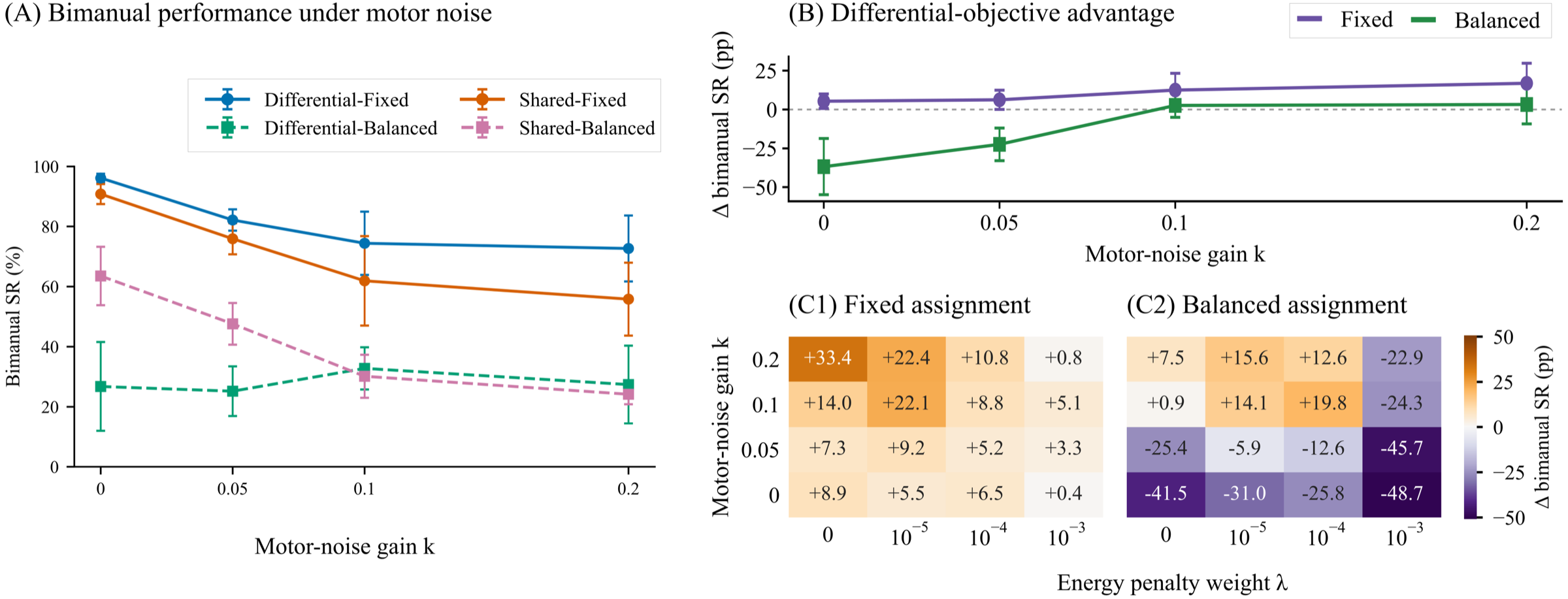}
  \caption{Bimanual performance under control constraints. \textbf{(A)} Success rate across motor-noise gain for four training conditions. \textbf{(B)} Differential--Shared Bimanual SR differences under fixed and balanced assignment. \textbf{(C)} Differential--Shared differences across noise and energy penalty under \textbf{(C1)} fixed and \textbf{(C2)} balanced assignment. Results average over undisplayed factors, with contrasts computed within matched conditions and seeds. A--B error bars show 95\% Student's $t$ confidence intervals across seeds. Positive differences favour Differential training.}
  \label{fig:constraints}
\end{figure}

Absolute performance shows that Differential-Fixed remains the highest-performing
training condition across all tested motor-noise levels
(Fig.~\ref{fig:constraints}A). As motor-noise gain increases, stronger signal-dependent perturbations make control more difficult, and Bimanual SR declines for most training conditions. Differential-Fixed nevertheless maintains comparatively high performance throughout the tested range. Shared-Fixed also remains substantially stronger than the balanced conditions, although its performance declines more sharply as motor-noise gain increases.

\vspace{-0.4em}

Given that Differential-Fixed achieved the best performance, we next examined how the performance of Differential and Shared training changes with motor-noise gain (Fig.~\ref{fig:constraints}B). Under fixed assignment, the Differential--Shared Bimanual SR advantage increased with motor-noise gain, reaching 17 percentage points at $k=0.2$ when averaged across energy penalties. Under balanced assignment, Differential training performed substantially worse than Shared training at low motor-noise gain, but this disadvantage narrowed and approached zero at the two highest gain levels. Thus, at higher motor-noise gain, the relative performance benefit of differential training became more pronounced under both task-assignment schemes, with a consistent advantage under fixed assignment.

Energy penalty further modulated this effect
(Fig.~\ref{fig:constraints}C). Differential training showed its largest
advantages under weaker energy penalties, particularly when motor noise was
high. Under fixed assignment, the largest difference reached 33.4 percentage
points at $\lambda=0$ and $k=0.2$, whereas the advantage became small at
stronger penalties. Under balanced assignment, Differential training exceeded
Shared training mainly under moderate-to-high motor noise and weak-to-moderate
energy penalties. Stronger energy penalties therefore reduced the additional
performance benefit of differential training objectives. Taken together, the relative performance of differential objectives was more favourable at higher motor-noise gain, particularly under fixed task assignment. The comparatively high performance of Differential-Fixed at larger motor-noise gains indicates greater tolerance to motor-command-dependent execution noise.

\subsection{Communication delay increases performance but non-monotonically alters specialisation}
\label{sec:delay}
\vspace{-0.2em}
Across four training conditions and control settings, Bimanual SR increased from 50.38\% at $d=0$ to 53.57\%, 56.32\%, and 61.55\% at $d=5$, $10$, and $20$ simulation steps (50, 100, and 200 ms), respectively. Functional specialisation showed a non-monotonic response to communication delay. Signed role alignment decreased from 0.0502 at $d=0$ to 0.0312 and 0.0217 at $d=5$ and $d=10$, before returning to 0.0416 at $d=20$. Functional separation showed a similar pattern, decreasing from 0.1533 to 0.1329 and 0.1280 before recovering to 0.1473 at $d=20$. The largest reduction in both specialisation measures occurred at the delay of $d=10$, whereas the longest delay produced values close to the no-delay condition. Because communication delay did not show a graded relationship with functional specialisation, it was treated as a categorical factor rather than a graded control axis in the trend analyses. The comparisons are reported in Appendix Table~\ref{tab:delay-level-results}.
\vspace{-0.2em}
\subsection{Role-Specific Performance--Flexibility Trade-Off}
\vspace{-0.2em}
To assess whether the performance advantage under trained hand--task roles transferred to altered task demands without retraining, we evaluated reversed hand--task mappings, symmetric bimanual tasks in which both hands performed tracking or holding, and perturbed-tracking tasks in which either hand tracked while resisting external perturbations (Fig.~\ref{fig:transfer}). Task definitions, evaluation procedures, and full results are provided in Appendix~\ref{app:alternative-evaluations}.

\begin{figure}[h]
  \centering
  \includegraphics[width=\linewidth]{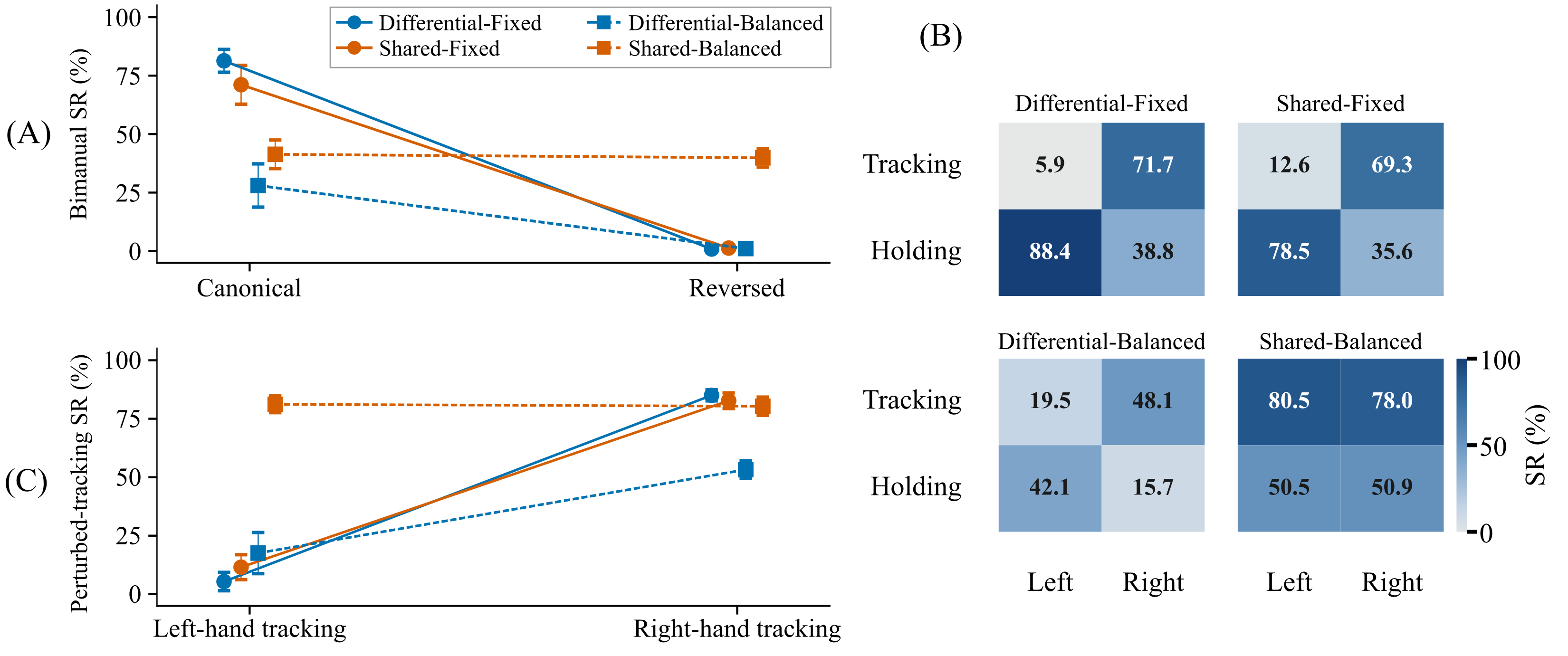}
  \caption{
    Hand-role flexibility across changing task demands without retraining.
    \textbf{(A)} Bimanual SR under canonical and reversed hand--task mappings.
    \textbf{(B)} Hand-specific tracking and holding success rates when both hands perform the same function.
    \textbf{(C)} Perturbed-tracking success rate when either hand tracks under external perturbation.
    Means are averaged over 64 control conditions. Panels A and C show 95\% Student's $t$ confidence intervals across five seeds.
    }
  \label{fig:transfer}
\end{figure}
\vspace{-0.2em}
Performance under altered hand--task roles differed substantially across training conditions (Fig.~\ref{fig:transfer}A--B). Differential-Fixed achieved 81.33\% Bimanual SR under the canonical mapping but only 0.87\% after role reversal, with tracking and holding competence concentrated in the trained hands. Shared-Fixed showed a similar dependence on trained hand roles. By comparison, Shared-Balanced retained similar Bimanual SR across canonical and reversed mappings (41.36\% and 39.85\%) and more even competence across the two hands. Differential-Balanced remained poor after role reversal, showing that balanced assignment alone was insufficient. Reversal reduced Bimanual SR by 80.46 percentage points for Differential-Fixed, compared with only 1.51 percentage points for Shared-Balanced. The same pattern appeared under perturbed tracking (Fig.~\ref{fig:transfer}C). Differential-Fixed achieved 85.01\% success when tracking was assigned to the trained hand but only 5.39\% with the opposite hand, whereas Shared-Balanced performed similarly with either hand (80.34\% and 81.17\%). The between-hand gap was therefore 79.62 percentage points for Differential-Fixed but only 0.83 percentage points for Shared-Balanced.

These results indicate that stronger role-specific specialisation was associated with higher performance when task demands matched the trained hand roles, but lower flexibility when those roles changed. Shared-Balanced training showed the opposite pattern, with lower canonical-task performance but greater flexibility across hands and task demands.
\vspace{-0.2em}

\section{Discussion and conclusion}
\vspace{-0.2em}

Our main finding is that functional specialisation, a division of labour in which different controllers contribute according to their respective strengths, was associated with behavioural advantages in bilateral robotic control. Differential-Fixed, which developed the strongest functional specialisation, achieved the highest bimanual success, with the highest tracking and holding success rates and the lowest errors, together with the lowest energy per success under the trained tasks. Further, this advantage remained evident under challenging control conditions. Higher motor-noise gain increases the magnitude of execution noise for the same motor command, reducing command-to-output reliability, which can result from component wear, abnormal friction, and joint backlash in robotic systems \citep{heInsituFaultDiagnosis2023,kimPhysicsguidedResidualLearning2026}. Beyond execution reliability, robotic systems face energy constraints, making both tolerance to unreliable motor execution and energy-efficient control important for maintaining performance and extending operational capacity. The model with the strongest functional specialisation, Differential-Fixed, remained the strongest performer as motor-noise gain increased, with its advantage becoming more favourable at higher noise levels, suggesting greater tolerance to unreliable motor execution. It also remained the strongest performer across energy penalties, although its advantage narrowed as energetic cost was weighted more strongly. 

This behaviourally advantageous functional specialisation depended on how learning responsibility was organised across controllers and hand roles within a system in which each controller has a stronger routing gain to one hand. Controller-specific functional objectives produced the clearest task-aligned specialisation when paired with stable hand--task assignment, as in Differential-Fixed. Differential-Balanced training makes this interaction clear: controller-specific objectives alone were insufficient. Under fixed assignment, each controller was repeatedly associated with the same function through its more strongly routed hand \citep{aracArmDominanceEmerges}, whereas balanced assignment varied these hand-specific roles and produced weaker task-aligned functional specialisation.

Biological motor systems balance functional specialisation with redundant control, allowing different components to develop complementary strengths while retaining some capacity to compensate when normal roles change \citep{serrienDynamicsHemisphericSpecialization2006a}. Our results reveal a similar trade-off in bilateral robotic control. Differential-Fixed achieved the strongest role-specific performance and efficiency, but its competence was concentrated in the trained hand--task roles, leading to a large loss of performance when those roles were reversed or reassigned. Shared-Balanced showed the opposite pattern: lower performance under the trained mapping, but much greater interchangeability across hands and altered task demands. Functional specialisation therefore affected not only internal controller organisation but also how behavioural competence was distributed across the system. For robotic control, this suggests that stronger specialisation may be advantageous when task roles are stable and repeatedly reused, whereas greater redundancy may be preferable when hand roles must change or when the system needs to compensate for altered operating demands.
\vspace{-0.2em}

\paragraph{Limitations and future work}
Our findings are limited to the bilateral motor-control setting and task family studied here. Biological principles are used to motivate the separation of controller-level functional specialisation and stable hand-role preference, but our model does not attempt to reproduce the biological mechanisms underlying human handedness or hemispheric specialisation. Hand-role flexibility is evaluated only through reassignment within this setting, leaving broader generalisation across task families and embodiments outside the scope. Because contralateral routing was fixed across all training conditions, the results characterise learning responsibility within a bilateral architecture in which each controller has a stronger routing gain to one hand and a weaker gain to the other. Alternative controller-to-arm mappings are not examined. Because Fixed and Balanced conditions differed in the hand--task mappings encountered during training, the observed flexibility differences reflect training exposure together with objective structure and learned organisation. More generally, because training structure affects both functional organisation and behaviour, these experiments do not isolate whether specialisation itself mediates the observed behavioural advantages independently of the training structure that produced it.

Future work could test broader task families, simulation environments, and physical robots, and compare with human behavioural or neural data if extending the biological interpretation.

\clearpage
\subsection*{AI use statement}
Generative AI tools were used to refine and improve wording based on author-written drafts, with all AI-assisted text manually reviewed and further edited by the authors. AI was also used to support literature retrieval, including identifying potentially relevant work for related-work discussions and methodological references, with all cited papers read and verified by the authors. All references were managed by the authors by importing the original PDFs into Zotero, exporting the corresponding BibTeX entries, and manually checking each entry. No reference entry was directly generated by AI. AI tools assisted with code generation and editing, and all resulting code was reviewed by the authors. They were also used for tooling tasks, such as generating editable source SVG files for architecture diagrams, all of which were subsequently manually modified. No directly AI-generated figures were used in the paper. The authors take full responsibility for the final content of this work.

\subsection*{Ethics statement}
This work consists entirely of computational simulations and does not involve
human participants, animal experiments, or the collection or use of personal
data. The controllers are evaluated in a simulated environment, with no
deployment on physical robots or testing involving people.

\subsection*{Reproducibility statement}
\phantomsection
\label{sec:reproducibility-statement}
Experiments used Python 3.10.19, PyTorch 2.5.1 with CUDA 12.1, NumPy 2.2.5,
and MotorNet 0.2.0. Models were trained on NVIDIA L40S-class GPUs. The
complete sweep comprised 1,280 training runs and required approximately 704
GPU-hours. Code is available at:
\href{https://github.com/jingwen-ava-li/code202609}{github.com/jingwen-ava-li/code202609}.

\clearpage
\bibliographystyle{arxiv_preprint}
\bibliography{references}

\clearpage
\appendix
\section{Experimental setup and training}
\label{app:experimental-setup}

\subsection{Musculoskeletal environment and task definitions}
\label{app:task}

Simulations used MotorNet 0.2.0 with the default parameters of RigidTendonArm26 and RigidTendonHillMuscle, comprising shoulder flexor/extensor, elbow flexor/extensor, and biarticular flexor/extensor muscles \citep{codolMotorNetPythonToolbox2024a}. The simulation step was
$\Delta t = 0.01$ s and each episode lasted $T=100$ steps. Shoulder
origins were separated by 0.4 m, with the left arm mirrored horizontally
in the workspace. Arms started from shoulder and elbow angles
of $45^\circ$ and $90^\circ$, respectively. 

For the tracking task, let $\mathbf r_0$ denote the initial endpoint of the
tracking hand. The target trajectory was

\begin{equation}
\mathbf r_t
=
\mathbf r_0
+
s a \sin(2\pi f t\Delta t)\mathbf e_y ,
\end{equation}

where $a\sim\mathcal U(0.04,0.06)$ m,
$f\sim\mathcal U(0.75,1.25)$ Hz, and
$s\in\{-1,+1\}$ was sampled with equal probability. The holding target
remained at the initial endpoint of the holding hand.

The holding hand received a constant Cartesian force pulse with onset sampled
uniformly from steps 5--10, duration from 10--15 steps, magnitude from
0.5--1.5 N, and direction uniformly from $[0,2\pi)$. The tracking hand
received no external force pulse in the main task.

\subsection{Controller architecture and control constraints}
\label{app:architecture}

Let $x_t$ denote the shared 46-dimensional sensory observation. It comprised proprioceptive muscle states, hand endpoint positions,
target-relative features, and bilateral spatial features describing the
relationship between the two hands. The recurrent
updates were

\begin{align}
h_t^L &=
\mathrm{GRU}_L([x_t,C_{R\to L}\bar h_t^R],h_{t-1}^L),\\
h_t^R &=
\mathrm{GRU}_R([x_t,C_{L\to R}\bar h_t^L],h_{t-1}^R).
\end{align}

Here, $C_{R\to L}$ and $C_{L\to R}$ are learned communication projections.
For a state from controller $j\in\{L,R\}$, the available communicated state was
$\bar h_t^j=h_{t-d}^j$ for $d>0$; for $d=0$, the latest causally available
state $\bar h_t^j=h_{t-1}^j$ was used. The tested positive delays of 5, 10,
and 20 simulation steps therefore corresponded to 50, 100, and 200 ms.

To implement the fixed 80:20 contralateral-to-ipsilateral routing, let
$p=0.8$, $c=\sqrt{2/(p^2+(1-p)^2)}$, and let $\mathbf 1_6$ denote a
six-dimensional vector of ones. With the first six motor components
corresponding to the left arm and the remaining six to the right arm,

\begin{equation}
g_L=c[(1-p)\mathbf1_6,p\mathbf1_6],\qquad
g_R=c[p\mathbf1_6,(1-p)\mathbf1_6].
\end{equation}

The deterministic motor command and its signal-dependent noisy execution were

\begin{align}
u_t &=
\sigma(g_L\odot W_Lh_t^L+g_R\odot W_Rh_t^R+b),\\
\widetilde u_t &=
\mathrm{clip}_{[0,1]}
\left(u_t\odot(1+k\epsilon_t)\right),
\qquad
\epsilon_t\sim\mathcal N(0,I_{12}),
\end{align}

with fixed motor bias $b=0.1$. The routing gains scaled the learned readout
outputs and therefore did not impose a fixed fraction of executed control from
either controller.

Communication delay $d$ and motor-noise gain $k$ acted directly on the forward
control process defined above, whereas the energy-penalty weight $\lambda$
entered the training objectives in Appendix~\ref{app:objectives}.

\subsection{Training procedure and objectives}
\label{app:objectives}
\label{app:training-protocol}

\paragraph{Training configuration.}
Models were trained with Adam \citep{kingmaAdamMethod2015} using a
learning rate of $10^{-3}$ and a batch size of 128. Each epoch comprised one
newly sampled batch of one-second episodes and one parameter update. Training
ran for at most 2,000 epochs. Five random seeds (0--4) were used for each
control condition. Training used validation-based early stopping, and the checkpoint with the
best validation performance over the hand--task mappings used during training
was retained for evaluation.

\paragraph{Loss definitions and gradient assignment.}

For trial $b\in\{1,\ldots,B\}$ and time step
$t\in\{1,\ldots,T\}$, let $e^T_{bt}$ and $e^H_{bt}$ denote the Euclidean
endpoint errors of the tracking and holding hands, respectively. Here, $B$
is the batch size and $T$ is the number of simulation steps per episode.
Both errors were penalised using the Huber loss $\rho_\delta(\cdot)$ with
transition $\delta=0.01$ m \citep{huberRobustEstimationLocation1964}.
Let $m_{bt}$ be an indicator equal to one from perturbation onset to the end
of the episode and zero otherwise, and let $u_{bt}\in\mathbb{R}^{12}$ denote
the deterministic motor command before signal-dependent noise and clipping.
The functional losses and motor-output penalty were

\begin{align}
L_T &=
\frac{1}{BT}\sum_{b,t}\rho_\delta(e^T_{bt}),\\
L_H &=
\frac{\sum_{b,t}m_{bt}\rho_\delta(e^H_{bt})}
     {\sum_{b,t}m_{bt}},\\
E &=
\frac{1}{BT}\sum_{b,t}\lVert u_{bt}\rVert_2^2.
\end{align}

Thus, tracking loss was evaluated over the full episode, whereas holding loss
was evaluated from perturbation onset through recovery. The energy term used
the deterministic motor commands before signal-dependent noise and clipping.

The controller-owned parameter groups were

\begin{equation}
\theta_L=
\{\mathrm{GRU}_L,W_L,C_{R\to L}\},
\qquad
\theta_R=
\{\mathrm{GRU}_R,W_R,C_{L\to R}\}.
\end{equation}

Under Differential training, the gradients were

\begin{equation}
g_{\theta_L}
=
\nabla_{\theta_L}(L_T+\lambda E),
\qquad
g_{\theta_R}
=
\nabla_{\theta_R}(L_H+\lambda E).
\end{equation}

\begin{table}[h]
  \centering
  \caption{Training implementation for the learning-responsibility
  conditions. Canonical denotes right-hand tracking and left-hand holding.
  Reversed denotes left-hand tracking and right-hand holding.}
  \label{tab:training-conditions}
  \begin{tabular}{lll}
  \toprule
  Condition & Batch composition & Training objective \\
  \midrule
  Differential-Fixed
  & 128 canonical
  & $\theta_L: L_T+\lambda E,\quad \theta_R: L_H+\lambda E$ \\
  
  Differential-Balanced
  & 64 canonical + 64 reversed
  & $\theta_L: L_T+\lambda E,\quad \theta_R: L_H+\lambda E$ \\
  
  Shared-Fixed
  & 128 canonical
  & $\theta_L,\theta_R: L_T+L_H+\lambda E$ \\
  
  Shared-Balanced
  & 64 canonical + 64 reversed
  & $\theta_L,\theta_R: L_T+L_H+\lambda E$ \\
  \bottomrule
  \end{tabular}
  \end{table}

Under Differential-Balanced training, controller-level functional
responsibilities remained fixed when the hand--task mapping was reversed.
Gradient assignment affected parameter updates only. Both controllers, arms,
and communication pathways remained active in the forward computation.

\section{Evaluation metrics}
\label{app:evaluation-metrics}

\subsection{Functional specialisation metrics}
\label{app:representation}

For the controlled functional-specialisation assessment, each trained model was evaluated under the canonical hand--task mapping without
parameter updates. Execution noise was set to zero to isolate responses to
controlled task demands. Tracking frequency took values
$\{0.75,1.0,1.25\}$ Hz and holding force magnitude took values
$\{0.5,1.0,1.5\}$ N, giving a $3\times3$ factorial design. Tracking amplitude
was fixed at 0.05 m, and each demand combination used 64 trials.

\paragraph{Unit-level task preference.}
For hidden unit $i$ in trial $b$, activity was summarised by its root mean
square over the episode,

\begin{equation}
a_{ib}=
\sqrt{\frac{1}{T}\sum_{t=1}^{T}h_{ibt}^2}.
\end{equation}

Within each evaluation stream, unit activity, tracking frequency, and holding
force were standardised. Ordinary least squares fitted

\begin{equation}
z(a_{ib})=
\beta_{0i}
+\beta_{Ti}z(f_b)
+\beta_{Hi}z(F_b)
+\beta_{THi}z(f_b)z(F_b)
+\varepsilon_{ib}.
\end{equation}

Adapting the fractional task-selectivity formulation of
\citet{yangTaskRepresentationsNeural2019}, relative preference between the two
functional demands was defined as

\begin{equation}
p_i=
\frac{|\beta_{Ti}|-|\beta_{Hi}|}
     {\max(|\beta_{Ti}|+|\beta_{Hi}|,10^{-12})}.
\end{equation}

Positive values indicate greater sensitivity to tracking frequency and
negative values greater sensitivity to holding force. All 64 units per
controller were included.

\paragraph{Signed role alignment.}
Let $\bar p_L$ and $\bar p_R$ denote the mean unit preferences of the left and
right controllers. Signed role alignment was

\begin{equation}
A_{\mathrm{role}}=
\frac{\bar p_L-\bar p_R}{2}.
\end{equation}

Controller identities were kept fixed when computing role alignment and were not swapped based on their learned preferences. Positive values indicate relatively greater
tracking preference in the left controller.

\paragraph{Functional separation.}
Functional separation was defined as the one-dimensional Wasserstein distance
between the two empirical preference distributions
\citep{peyreComputationalOptimalTransport2019}:

\begin{equation}
D_W=
\frac{1}{64}
\sum_{i=1}^{64}
|p^L_{(i)}-p^R_{(i)}|,
\end{equation}

where parentheses indicate separately sorted unit preferences. Unlike signed
role alignment, this measure is invariant to exchanging controller labels and
quantifies the magnitude of between-controller differentiation irrespective of
direction.

These two functional-specialisation metrics quantify relative sensitivity to
the manipulated tracking and holding demands under the canonical hand--task
mapping.

\subsection{Behavioural performance metrics}
\label{app:behaviour}

For the performance assessment, canonical and reversed mappings were each
evaluated over 128 trials sampled from the same task distributions as training.
Model parameters were frozen, and execution-noise gain was set to the model's training value $k$. For each control condition and training seed, model families were evaluated on the same set of sampled trials.

\paragraph{Tracking and holding errors.}
For trial $b$, the whole-episode mean tracking error and the
pulse-and-recovery mean holding error were

\begin{align}
\overline e_b^T
&=
\frac{1}{T}\sum_t e^T_{bt},\\
\overline e_b^H
&=
\frac{\sum_t m_{bt}e^H_{bt}}
     {\sum_t m_{bt}},
\end{align}

where $m_{bt}$ selects time steps from perturbation onset through recovery.
Reported trajectory and holding errors were the corresponding means across
$N$ evaluation trials:

\begin{equation}
\mathrm{Err}_T=\frac{1}{N}\sum_{b=1}^{N}\overline e_b^T,
\qquad
\mathrm{Err}_H=\frac{1}{N}\sum_{b=1}^{N}\overline e_b^H.
\end{equation}

\paragraph{Success rates.}
Let $\mathbb I[\cdot]$ denote the indicator function. Tracking SR, Holding SR,
and Bimanual SR were

\begin{align}
S_T
&=
\frac{1}{N}\sum_{b=1}^{N}
\mathbb I[\overline e_b^T\leq0.02],\\
S_H
&=
\frac{1}{N}\sum_{b=1}^{N}
\mathbb I[\overline e_b^H\leq0.01],\\
S_B
&=
\frac{1}{N}\sum_{b=1}^{N}
\mathbb I[
\overline e_b^T\leq0.02
\land
\overline e_b^H\leq0.01].
\end{align}

Bimanual SR therefore required both task criteria to be satisfied in the same
trial. For symmetric evaluations, it analogously required both hands to satisfy
the relevant tracking or holding criterion in the same trial. All success
criteria were based on mean error.

\subsection{Control efficiency metrics}
\label{app:control-efficiency}

\paragraph{Mean control energy.}
Mean control energy was computed from the deterministic motor commands
$u_{bt}$ defined in Appendix~\ref{app:objectives}, before signal-dependent noise
and clipping, and included both successful and failed trials:

\begin{equation}
\overline E=
\frac{1}{NT}
\sum_{b=1}^{N}\sum_{t=1}^{T}
\lVert u_{bt}\rVert_2^2.
\end{equation}

\paragraph{Energy per success.}
Energy per success was defined as $\overline E/S_B$, where the numerator and
denominator were averaged over the same analysis group before forming the
ratio. It therefore represents control expenditure relative to success rate.

\section{Main task results and statistical analysis}
\label{app:statistics}

\subsection{Full-grid performance and functional organisation}
\label{app:specialisation-results}

Table~\ref{tab:organisation-summary} summarises all 64 constraint cells for
each training condition. The cells comprise four communication delays, four
energy-penalty weights, and four motor-noise gains. Each entry was calculated
by first averaging the 64 cells within each training seed and then summarising
the five resulting seed means. Bimanual SR uses each model's native
execution-noise gain. Functional separation and signed role alignment use the
noise-free controlled-demand evaluation defined in
Appendix~\ref{app:representation}.

\begin{table}[htbp]
 \centering
 \small
 \caption{Performance and controller organisation across the complete
 constraint grid. Values are mean [95\% confidence interval] across five
 seed-level grid means. Each seed-level value averages the same 64 constraint
 settings. Bimanual SR is reported as a percentage.}
 \label{tab:organisation-summary}
 \begin{tabular}{@{}lccc@{}}
 \toprule
 Training condition & Bimanual SR & Signed role alignment & Functional separation \\
 \midrule
Differential-Fixed & $81.33\ [76.45, 86.20]$ & $0.0811\ [0.0719, 0.0904]$ & $0.190\ [0.177, 0.203]$ \\
Shared-Fixed & $71.10\ [62.80, 79.40]$ & $0.0292\ [0.0201, 0.0384]$ & $0.120\ [0.110, 0.130]$ \\
Differential-Balanced & $28.03\ [18.78, 37.28]$ & $-0.0004\ [-0.0076, 0.0068]$ & $0.133\ [0.119, 0.147]$ \\
Shared-Balanced & $41.36\ [35.25, 47.47]$ & $0.0347\ [0.0204, 0.0490]$ & $0.118\ [0.101, 0.135]$ \\
 \bottomrule
 \end{tabular}
\end{table}

\paragraph{Objective-by-assignment interactions.}
For each organisation metric, the interaction contrast was computed within
each seed after averaging over the 64 constraint settings:
$(\mathrm{Differential\!-\!Fixed}-\mathrm{Shared\!-\!Fixed})
-(\mathrm{Differential\!-\!Balanced}-\mathrm{Shared\!-\!Balanced})$.
The five seed-level contrasts were tested against zero using two-sided
one-sample $t$ tests ($df=4$). Holm correction was applied jointly across the
two organisation metrics. Both interactions remained significant after
correction (signed role alignment: $p_{\mathrm H}=3.96\times10^{-5}$;
functional separation: $p_{\mathrm H}=0.00295$).

Functional separation is non-negative and label-invariant, whereas signed role
alignment preserves the anatomical controller ordering. A larger distribution
distance can therefore coexist with little directional role alignment.

\paragraph{Supplementary controller lesion analysis}
 \label{app:lesion-analysis}
We performed lesions as evaluation-time interventions on trained models without retraining or modifying learned weights. For the lesioned controller, both its outgoing communication signal and motor contribution were multiplied by a retained gain $\gamma$. For the left controller, $\tilde{\mathbf z}_{L\rightarrow R,t}=\gamma_L C_{L\rightarrow R}\bar{\mathbf h}^L_t$ and $\tilde{\mathbf c}_{L,t}=\gamma_L\left(\mathbf g_L\odot W_L\mathbf h^L_t\right)$, with the corresponding operation applied to the right controller. A gain of $\gamma=1$ denotes the intact model and $\gamma=0$ removes all outgoing influence from the lesioned controller while leaving its internal recurrent state and incoming signals unchanged. We therefore refer to $\gamma=0$ as a whole-module functional lesion.

Lesions were evaluated under the canonical right-tracking/left-holding mapping using $\gamma\in\{0,0.95,0.975,0.99\}$. For each checkpoint, intact, left-lesion, and right-lesion evaluations used matched trials and motor-noise realisations. Within each seed, outcomes were averaged equally over the 64 delay--energy--noise conditions and then summarised across five seeds. We report changes relative to the intact model as $\Delta=\mathrm{lesion}-\mathrm{intact}$.

\begin{table*}[h]
\centering
\caption{Graded controller lesion effects on canonical-task success rates. Values are mean $\pm$ SD across five seed-level means, with each seed averaged equally over the 64 control settings. Changes are calculated as lesion minus intact in percentage points. Retained gain $\gamma$ scales the lesioned controller's outgoing influence, and negative values indicate reduced success after lesion.}
\label{tab:graded-lesion-success}
\setlength{\tabcolsep}{3.2pt}
\renewcommand{\arraystretch}{1.15}
\resizebox{\textwidth}{!}{%
\begin{tabular}{llrrrrrrrr}
\toprule
& & \multicolumn{4}{c}{$\Delta$ Tracking SR (pp)}
& \multicolumn{4}{c}{$\Delta$ Holding SR (pp)} \\
\cmidrule(lr){3-6}
\cmidrule(lr){7-10}
Model & Lesioned controller
& $\gamma=0$ & $\gamma=0.95$ & $\gamma=0.975$ & $\gamma=0.99$
& $\gamma=0$ & $\gamma=0.95$ & $\gamma=0.975$ & $\gamma=0.99$ \\
\midrule

Differential-Fixed
& Left
& $-92.31 \pm 1.88$
& $-19.87 \pm 8.33$
& $-6.17 \pm 3.01$
& $-0.89 \pm 0.91$
& $-86.13 \pm 4.26$
& $-7.38 \pm 1.36$
& $-1.43 \pm 0.71$
& $-0.30 \pm 0.19$ \\
& Right
& $-91.57 \pm 1.70$
& $-10.78 \pm 6.14$
& $-3.41 \pm 2.77$
& $-0.24 \pm 0.23$
& $-87.08 \pm 4.67$
& $-19.94 \pm 2.62$
& $-5.74 \pm 1.67$
& $-0.97 \pm 0.63$ \\
\midrule

Shared-Fixed
& Left
& $-89.28 \pm 2.35$
& $-6.14 \pm 2.84$
& $-0.92 \pm 0.71$
& $-0.12 \pm 0.32$
& $-66.03 \pm 4.67$
& $-0.14 \pm 0.24$
& $+0.02 \pm 0.23$
& $+0.08 \pm 0.19$ \\
& Right
& $-86.33 \pm 2.83$
& $-0.48 \pm 0.45$
& $-0.05 \pm 0.25$
& $+0.08 \pm 0.16$
& $-76.77 \pm 7.81$
& $-5.59 \pm 1.36$
& $-1.63 \pm 0.43$
& $-0.45 \pm 0.25$ \\
\midrule

Differential-Balanced
& Left
& $-64.33 \pm 4.17$
& $-40.68 \pm 7.19$
& $-26.06 \pm 3.91$
& $-10.50 \pm 1.97$
& $-45.05 \pm 10.03$
& $-36.38 \pm 9.11$
& $-27.51 \pm 8.24$
& $-12.93 \pm 3.16$ \\
& Right
& $-64.33 \pm 4.17$
& $-37.72 \pm 8.99$
& $-23.25 \pm 4.44$
& $-7.68 \pm 0.92$
& $-45.05 \pm 10.03$
& $-41.62 \pm 9.36$
& $-34.99 \pm 9.75$
& $-19.78 \pm 4.87$ \\
\midrule

Shared-Balanced
& Left
& $-84.38 \pm 2.16$
& $-3.25 \pm 1.13$
& $-0.43 \pm 0.40$
& $+0.10 \pm 0.22$
& $-45.53 \pm 6.05$
& $-0.68 \pm 0.93$
& $-0.18 \pm 0.53$
& $-0.08 \pm 0.24$ \\
& Right
& $-76.86 \pm 5.54$
& $-0.68 \pm 0.45$
& $-0.21 \pm 0.24$
& $-0.07 \pm 0.07$
& $-46.82 \pm 5.69$
& $-9.09 \pm 3.38$
& $-3.60 \pm 1.67$
& $-0.92 \pm 0.47$ \\

\bottomrule
\end{tabular}%
}
\end{table*}

Complete lesions ($\gamma=0$) drove both task components to near-floor performance, showing that successful bimanual control depended on contributions from both controllers. Under partial attenuation, the direction of impairment was more selective: left-controller attenuation tended to reduce tracking more strongly, whereas right-controller attenuation reduced holding more strongly. Because controller identity is coupled to the primarily controlled hand by the fixed routing architecture, these directional effects reflect both controller--hand organisation and functional dependence. Lesion magnitude was therefore not treated as an independent measure of functional specialisation.

\subsection{Task performance comparisons}
\label{app:performance-efficiency-results}

The primary inferential question was whether differences between training
conditions were consistently expressed across the 64 prespecified constraint
settings. Accordingly, the five training seeds were first averaged within each constraint
cell to obtain one estimate per training condition and setting, and the
resulting 64 settings were treated as matched experimental conditions.
Differential-Fixed was then compared with each alternative across the resulting
64 matched settings using two-sided paired Monte Carlo permutation tests.

Under the null hypothesis, condition labels were exchanged within each matched
constraint setting. The test statistic was the difference between the two
conditions after equal weighting of the 64 settings. Thus, the analysis tests
for a systematic performance difference across the fixed constraint grid.
For energy per success, the ratio was recomputed from the corresponding
permuted energy and success-rate quantities after each label exchange.

Each comparison used $10^6$ permutations with random seed 20260918. Monte Carlo
probabilities used the plus-one correction
$p=(1+N_{\mathrm{extreme}})/(10^6+1)$
\citep{phipsonPermutationPvaluesShould2010}. Holm correction was applied jointly
to all 21 comparisons across seven metrics and three alternatives
\citep{holmSimpleSequentiallyRejective1979}. Table~\ref{tab:performance-tests} reports the Holm-adjusted results for all 21
canonical performance comparisons.

\begin{table}[htbp]
 \centering
 \small
 \caption{Holm-adjusted p-values from paired Monte Carlo permutation tests for Differential-Fixed versus
 each alternative across the 64 prespecified constraint settings. Tests assess
 systematic condition differences across the fixed constraint grid and are
 two-sided. Correction was applied jointly across all 21 comparisons.}
 \label{tab:performance-tests}
 \begin{tabular}{@{}lccc@{}}
 \toprule
 Metric & Shared-Fixed & Differential-Balanced & Shared-Balanced \\
 \midrule
Bimanual SR & $<0.001$ & $<0.001$ & $<0.001$ \\
Tracking SR & $0.001$ & $<0.001$ & $<0.001$ \\
Holding SR & $<0.001$ & $<0.001$ & $<0.001$ \\
Trajectory error & $0.004$ & $<0.001$ & $<0.001$ \\
Holding error & $<0.001$ & $<0.001$ & $<0.001$ \\
Mean control energy & $<0.001$ & $0.162$ & $<0.001$ \\
Energy per success & $<0.001$ & $<0.001$ & $<0.001$ \\
 \bottomrule
 \end{tabular}
\end{table}

The mean-control-energy difference between Differential-Fixed and
Differential-Balanced was not significant after correction
($p_{\mathrm{Holm}}=0.162$). All other comparisons remained significant after Holm correction.

\subsection{Communication-delay results}
\label{app:delay-results}

To estimate the effect of communication delay, outcomes were averaged within each seed over training condition, energy penalty, and motor-noise gain separately for each delay level. Reported means were then calculated across five seeds. Each nonzero delay was compared with $d=0$ using two-sided paired $t$ tests for Bimanual SR, signed role alignment, and functional separation. Confidence intervals were calculated across five paired seed-level differences, and Holm correction was applied across the nine comparisons. Table~\ref{tab:delay-level-results} reports the resulting comparisons.

\begin{table}[htbp]
    \centering
    \small
    \caption{Performance and functional specialisation by communication-delay level. Differences compare each nonzero delay with $d=0$. Bimanual SR differences are reported in percentage points. $p_{\mathrm{H}}$ denotes Holm-adjusted $p$-values.}
    \label{tab:delay-level-results}
    \begin{tabular}{@{}llccc@{}}
        \toprule
        Metric
        & Delay $d$
        & Mean
        & Difference from $d=0$ [95\% CI]
        & $p_{\mathrm{H}}$ \\
        \midrule

        Bimanual SR (\%)
        & 0  & 50.38 & Reference                         & --    \\
        & 5  & 53.57 & $+3.19$ [$-2.86$, $9.24$]        & 0.434 \\
        & 10 & 56.32 & $+5.94$ [$4.55$, $7.32$]         & 0.003 \\
        & 20 & 61.55 & $+11.16$ [$5.53$, $16.80$]       & 0.037 \\

        \addlinespace
        Signed role alignment
        & 0  & 0.0502 & Reference                         & --    \\
        & 5  & 0.0312 & $-0.0190$ [$-0.0337$, $-0.0043$] & 0.116 \\
        & 10 & 0.0217 & $-0.0285$ [$-0.0397$, $-0.0173$] & 0.017 \\
        & 20 & 0.0416 & $-0.0086$ [$-0.0217$, $0.0045$]  & 0.427 \\

        \addlinespace
        Functional separation
        & 0  & 0.1533 & Reference                         & --    \\
        & 5  & 0.1329 & $-0.0204$ [$-0.0390$, $-0.0017$] & 0.156 \\
        & 10 & 0.1280 & $-0.0253$ [$-0.0393$, $-0.0114$] & 0.044 \\
        & 20 & 0.1473 & $-0.0060$ [$-0.0189$, $0.0070$]  & 0.434 \\

        \bottomrule
    \end{tabular}
\end{table}

To examine whether the relative performance of Differential and Shared training varied across control regimes,
Table~\ref{tab:delay-contrasts} separates the Differential $-$ Shared Bimanual
SR contrasts by communication delay, motor-noise gain, and task assignment.
Within each combination, the four energy-penalty settings were averaged within
each seed. The stratified contrasts varied substantially across control regimes and did not show a consistent monotonic relationship with communication delay.

\begin{table}[htbp]
    \centering
    \small
    \caption{Delay-stratified Differential $-$ Shared Bimanual SR contrasts in
    percentage points.}
    \label{tab:delay-contrasts}
    \begin{tabular}{@{}rrcc@{}}
        \toprule
        Delay & Noise & Fixed assignment & Balanced assignment \\
        (steps) & $k$ & $\Delta$SR [95\% CI] & $\Delta$SR [95\% CI] \\
        \midrule
        0  & 0    & $6.8$ [$1.0$, $12.5$]    & $-30.4$ [$-54.6$, $-6.1$] \\
        0  & 0.05 & $13.0$ [$-11.0$, $37.0$] & $-22.1$ [$-45.1$, $0.9$] \\
        0  & 0.1  & $16.5$ [$9.3$, $23.8$]   & $-5.1$ [$-17.5$, $7.2$] \\
        0  & 0.2  & $20.4$ [$4.7$, $36.2$]   & $4.3$ [$-18.3$, $26.9$] \\
        \addlinespace
        5  & 0    & $1.6$ [$-0.8$, $4.1$]    & $-29.7$ [$-52.2$, $-7.2$] \\
        5  & 0.05 & $2.1$ [$-7.8$, $11.9$]   & $-19.2$ [$-34.7$, $-3.7$] \\
        5  & 0.1  & $15.4$ [$4.0$, $26.8$]   & $7.5$ [$-6.4$, $21.3$] \\
        5  & 0.2  & $30.7$ [$4.7$, $56.8$]   & $11.1$ [$-5.6$, $27.7$] \\
        \addlinespace
        10 & 0    & $5.2$ [$-5.0$, $15.4$]   & $-38.9$ [$-52.8$, $-25.1$] \\
        10 & 0.05 & $7.2$ [$-2.9$, $17.4$]   & $-22.5$ [$-35.5$, $-9.6$] \\
        10 & 0.1  & $4.6$ [$-7.0$, $16.3$]   & $3.3$ [$-10.6$, $17.2$] \\
        10 & 0.2  & $12.0$ [$-5.4$, $29.3$]  & $0.9$ [$-8.9$, $10.6$] \\
        \addlinespace
        20 & 0    & $7.7$ [$-5.4$, $20.9$]   & $-48.0$ [$-74.5$, $-21.4$] \\
        20 & 0.05 & $2.7$ [$-9.9$, $15.2$]   & $-25.8$ [$-32.8$, $-18.8$] \\
        20 & 0.1  & $13.4$ [$-4.5$, $31.3$]  & $4.9$ [$-9.9$, $19.7$] \\
        20 & 0.2  & $4.3$ [$-7.4$, $16.0$]   & $-3.3$ [$-17.2$, $10.6$] \\
        \bottomrule
    \end{tabular}
\end{table}

\section{Hand-role flexibility across task configurations}
\label{app:alternative-evaluations}

\subsection{Task definitions and evaluation protocol}
\label{app:additional-tasks}

Six task configurations were used to assess hand-role flexibility without parameter updates. M1 used the canonical mapping from the main task, whereas M2 reversed
the hand roles. S1 and S2 tested symmetric tracking and holding demands, while P1 and P2
tested perturbed tracking with either hand active, as defined in
Table~\ref{tab:alternative-protocol}.

Unless otherwise specified, target and perturbation distributions followed
Appendix~A.1. In S1, target amplitude, frequency, and movement direction were
sampled independently for the two hands. In S2, pulse directions were sampled
independently, while pulse magnitude and timing were shared within each trial.
In P1 and P2, the inactive hand retained a stationary target without a pulse
and was excluded from the success criterion. Each evaluation used 128 trials. For each evaluation and control condition,
the four training conditions were assessed on the same sampled trials. Model
parameters remained frozen, and execution-noise gain was set to the model's
training value $k$.

\begin{table}[H]
    \centering
    \small
    \caption{Definitions of the six hand--task configurations.
    Success criteria follow the behavioural metrics defined in
    Appendix~\ref{app:behaviour}.}
    \label{tab:alternative-protocol}
    \begin{tabular}{@{}p{0.08\linewidth}p{0.31\linewidth}p{0.54\linewidth}@{}}
        \toprule
        ID & Task demands & Success criterion \\
        \midrule
        M1 & Right hand tracks, left hand holds under a pulse (main task) &
        Tracking whole-episode mean error $\leq20$ mm and holding
        pulse-and-recovery mean error $\leq10$ mm in the same trial \\

        M2 & Left hand tracks, right hand holds under a pulse &
        Tracking whole-episode mean error $\leq20$ mm and holding
        pulse-and-recovery mean error $\leq10$ mm in the same trial \\

        S1 & Both hands track independent moving targets &
        Both hands' whole-episode mean tracking errors $\leq20$ mm \\

        S2 & Both hands hold fixed targets under external pulses &
        Both hands' pulse-and-recovery mean holding errors $\leq10$ mm \\

        P1 & Left hand tracks and receives a pulse &
        Left-hand mean tracking error $\leq20$ mm over both the full episode
        and pulse-and-recovery interval \\

        P2 & Right hand tracks and receives a pulse &
        Right-hand mean tracking error $\leq20$ mm over both the full episode
        and pulse-and-recovery interval \\
        \bottomrule
    \end{tabular}
\end{table}

\subsection{Performance across hand--task configurations}
\label{app:alternative-results}

For each evaluation, the 64 constraint settings were first averaged within
each training seed. Table~\ref{tab:alternative-complete} reports the mean and
95\% Student's $t$ confidence interval across the five resulting seed-level
grid means.

\begin{table}[htbp]
    \centering
    \scriptsize
    \caption{Performance across the six hand--task configurations. Values are
    success rates in percent, reported as mean [95\% confidence interval]
    across five seed-level grid means.}
    \label{tab:alternative-complete}

    \resizebox{\textwidth}{!}{%
    \begin{tabular}{@{}llcccc@{}}
        \toprule
        ID & Metric &
        \shortstack{Differential-\\Fixed} &
        \shortstack{Shared-\\Fixed} &
        \shortstack{Differential-\\Balanced} &
        \shortstack{Shared-\\Balanced} \\
        \midrule

        M1 & Bimanual SR &
        $81.33\ [76.45, 86.20]$ &
        $71.10\ [62.80, 79.40]$ &
        $28.03\ [18.78, 37.28]$ &
        $41.36\ [35.25, 47.47]$ \\

        M2 & Bimanual SR &
        $0.87\ [-0.41, 2.14]$ &
        $1.28\ [0.34, 2.22]$ &
        $1.07\ [0.70, 1.45]$ &
        $39.85\ [35.87, 43.82]$ \\

        \addlinespace

        S1 & Bimanual SR &
        $4.89\ [1.39, 8.39]$ &
        $9.78\ [5.21, 14.34]$ &
        $11.65\ [5.90, 17.41]$ &
        $63.52\ [58.81, 68.22]$ \\

        S1 & Left-hand SR &
        $5.94\ [1.72, 10.17]$ &
        $12.65\ [7.07, 18.23]$ &
        $19.51\ [11.28, 27.74]$ &
        $80.48\ [76.27, 84.68]$ \\

        S1 & Right-hand SR &
        $71.68\ [58.84, 84.53]$ &
        $69.34\ [62.89, 75.80]$ &
        $48.11\ [44.02, 52.21]$ &
        $78.01\ [74.43, 81.59]$ \\

        \addlinespace

        S2 & Bimanual SR &
        $36.67\ [25.80, 47.54]$ &
        $32.59\ [29.04, 36.13]$ &
        $9.70\ [6.68, 12.71]$ &
        $36.05\ [27.99, 44.11]$ \\

        S2 & Left-hand SR &
        $88.44\ [83.05, 93.82]$ &
        $78.50\ [68.88, 88.12]$ &
        $42.10\ [30.91, 53.30]$ &
        $50.52\ [41.96, 59.08]$ \\

        S2 & Right-hand SR &
        $38.81\ [26.42, 51.19]$ &
        $35.55\ [29.80, 41.31]$ &
        $15.70\ [12.06, 19.33]$ &
        $50.86\ [45.02, 56.70]$ \\

        \addlinespace

        P1 & Perturbed-tracking SR &
        $5.39\ [1.46, 9.31]$ &
        $11.53\ [6.19, 16.87]$ &
        $17.57\ [8.77, 26.36]$ &
        $81.17\ [77.61, 84.74]$ \\

        P2 & Perturbed-tracking SR &
        $85.01\ [82.65, 87.38]$ &
        $82.71\ [79.36, 86.05]$ &
        $53.26\ [49.47, 57.04]$ &
        $80.34\ [76.45, 84.23]$ \\

        \bottomrule
    \end{tabular}%
    }
\end{table}

The hand-specific rates clarify the source of performance differences in the
symmetric tasks. The hand-specific rates distinguish bilateral task success from competence of
the individual hands. In particular, low Bimanual SR can coexist with high
success in one hand when performance is strongly asymmetric across hands.

\subsection{Statistical tests of hand-role asymmetry}
\label{app:hand-role-asymmetry}

Four contrasts were used to quantify hand-role asymmetry within
each training condition: (1) M1 $-$ M2, (2) right $-$ left tracking success in S1,
(3) left $-$ right holding success in S2, and (4) P2 $-$ P1. Positive values indicate
better performance in the canonical right-tracking/left-holding direction. For each contrast, differences were first calculated within matched constraint
settings and training seeds, then averaged over the 64 settings within each
seed. The five resulting seed-level contrasts were tested against zero using
two-sided one-sample $t$ tests. Reported confidence intervals are 95\%
Student's $t$ intervals across the five seed-level contrasts. Holm correction
was applied jointly across the 16 tests.

\begin{table}[htbp]
    \centering
    \small
    \caption{Within-condition hand-role asymmetry contrasts. Differences are
    reported in percentage points as mean [95\% confidence interval].
    $p_{\mathrm H}$ denotes the Holm-adjusted $p$-value across the 16
    prespecified tests.}
    \label{tab:hand-role-tests}

    \begin{tabular}{@{}llcc@{}}
        \toprule
        Contrast & Training condition &
        Difference [95\% CI] & $p_{\mathrm H}$ \\
        \midrule

        M1 $-$ M2
        & Differential-Fixed
        & $80.46$ [$75.56$, $85.36$]
        & $<0.001$ \\

        & Shared-Fixed
        & $69.82$ [$60.83$, $78.80$]
        & $<0.001$ \\

        & Differential-Balanced
        & $26.96$ [$17.67$, $36.25$]
        & $0.009$ \\

        & Shared-Balanced
        & $1.51$ [$-3.29$, $6.32$]
        & $1.000$ \\

        \addlinespace

        S1 right $-$ left
        & Differential-Fixed
        & $65.74$ [$52.42$, $79.06$]
        & $0.002$ \\

        & Shared-Fixed
        & $56.69$ [$47.25$, $66.13$]
        & $<0.001$ \\

        & Differential-Balanced
        & $28.60$ [$20.94$, $36.27$]
        & $0.004$ \\

        & Shared-Balanced
        & $-2.47$ [$-7.57$, $2.64$]
        & $1.000$ \\

        \addlinespace

        S2 left $-$ right
        & Differential-Fixed
        & $49.63$ [$36.64$, $62.62$]
        & $0.004$ \\

        & Shared-Fixed
        & $42.95$ [$28.17$, $57.72$]
        & $0.009$ \\

        & Differential-Balanced
        & $26.41$ [$15.98$, $36.84$]
        & $0.011$ \\

        & Shared-Balanced
        & $-0.34$ [$-9.16$, $8.47$]
        & $1.000$ \\

        \addlinespace

        P2 $-$ P1
        & Differential-Fixed
        & $79.62$ [$75.16$, $84.09$]
        & $<0.001$ \\

        & Shared-Fixed
        & $71.18$ [$63.80$, $78.55$]
        & $<0.001$ \\

        & Differential-Balanced
        & $35.69$ [$25.81$, $45.57$]
        & $0.004$ \\

        & Shared-Balanced
        & $-0.83$ [$-5.27$, $3.61$]
        & $1.000$ \\

        \bottomrule
    \end{tabular}
\end{table}

Table~\ref{tab:hand-role-tests} shows that Differential-Fixed, Shared-Fixed,
and Differential-Balanced exhibited significant directional hand-role
asymmetries across all four contrasts after Holm correction. The largest
asymmetries were observed in the Fixed conditions, particularly
Differential-Fixed, indicating strong dependence on the canonical
right-tracking/left-holding allocation. In contrast, none of the
Shared-Balanced contrasts was significant, and all four differences were close
to zero. This pattern is consistent with substantially greater hand-role
flexibility under Shared-Balanced training.

\end{document}